\documentclass[letterpaper, 10 pt, conference]{ieeeconf}  

\IEEEoverridecommandlockouts                              

\usepackage{graphics} 
\usepackage{epsfig} 
\usepackage{times} 
\usepackage{amsmath} 
\usepackage{amssymb}  
\usepackage{algorithm}
\DeclareMathOperator*{\argmax}{arg\,max}

\usepackage[flushleft]{threeparttable}
\usepackage{multirow}
\usepackage[labelfont=bf, font=footnotesize]{caption}
\usepackage{subcaption}
\usepackage{booktabs}
\usepackage{pifont} 
\usepackage{amssymb}

\usepackage{amssymb}
\usepackage{graphicx}
\usepackage{amsmath}

\usepackage{rotating}
\usepackage{makeidx}
\usepackage{epstopdf}
\usepackage{xcolor}
\usepackage{colortbl}
\usepackage{color}
\usepackage{algpseudocode}

\makeatletter
\let\NAT@parse\undefined
\makeatother
\usepackage[pagebackref,breaklinks,colorlinks]{hyperref}

\title{\LARGE \bf
SAM-guided Pseudo Label Enhancement for Multi-modal 3D Semantic Segmentation
}

\author{Mingyu Yang, Jitong Lu, and Hun-Seok Kim 
\thanks{*This work was supported in part by COGNISENSE, one of seven centers in JUMP 2.0, a Semiconductor Research Corporation (SRC) program sponsored by DARPA.}
\thanks{Mingyu Yang, Jitong Lu, and Hun-Seok Kim are with Department of Electrical and Computer Engineering,
        University of Michigan, Ann Arbor, USA, 48109
        {\tt\small mingyuy@umich.edu, jitonglu@umich.edu, hunseok@umich.edu}}%
}

\begin{document}

\maketitle
\thispagestyle{empty}
\pagestyle{empty}

\newcommand{\figFront}{
\begin{figure}[t]
\centering
\includegraphics[width=0.95\columnwidth]{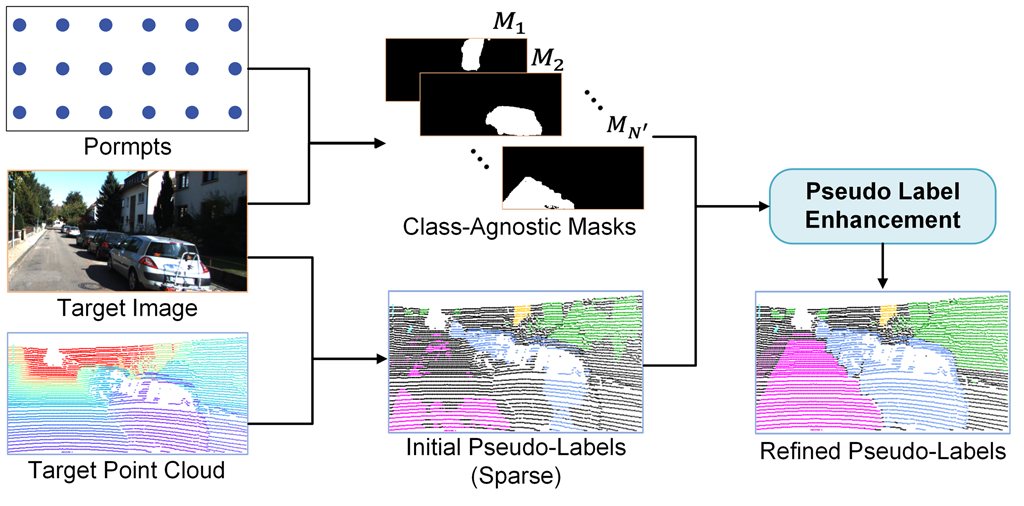}
\vspace{-0.05in}
\caption{\textbf{Overview of the proposed method.} The initial pseudo-labels tend to be accurate but sparse. Our proposed method aims to enhance the pseudo-labels by utilizing the grouping information of SAM masks.}
\label{fig:front}
\vspace{-0.3in}
\end{figure}
}

\newcommand{\figStructure}{
\begin{figure*}[t]
\centering
\includegraphics[width=0.9\linewidth]{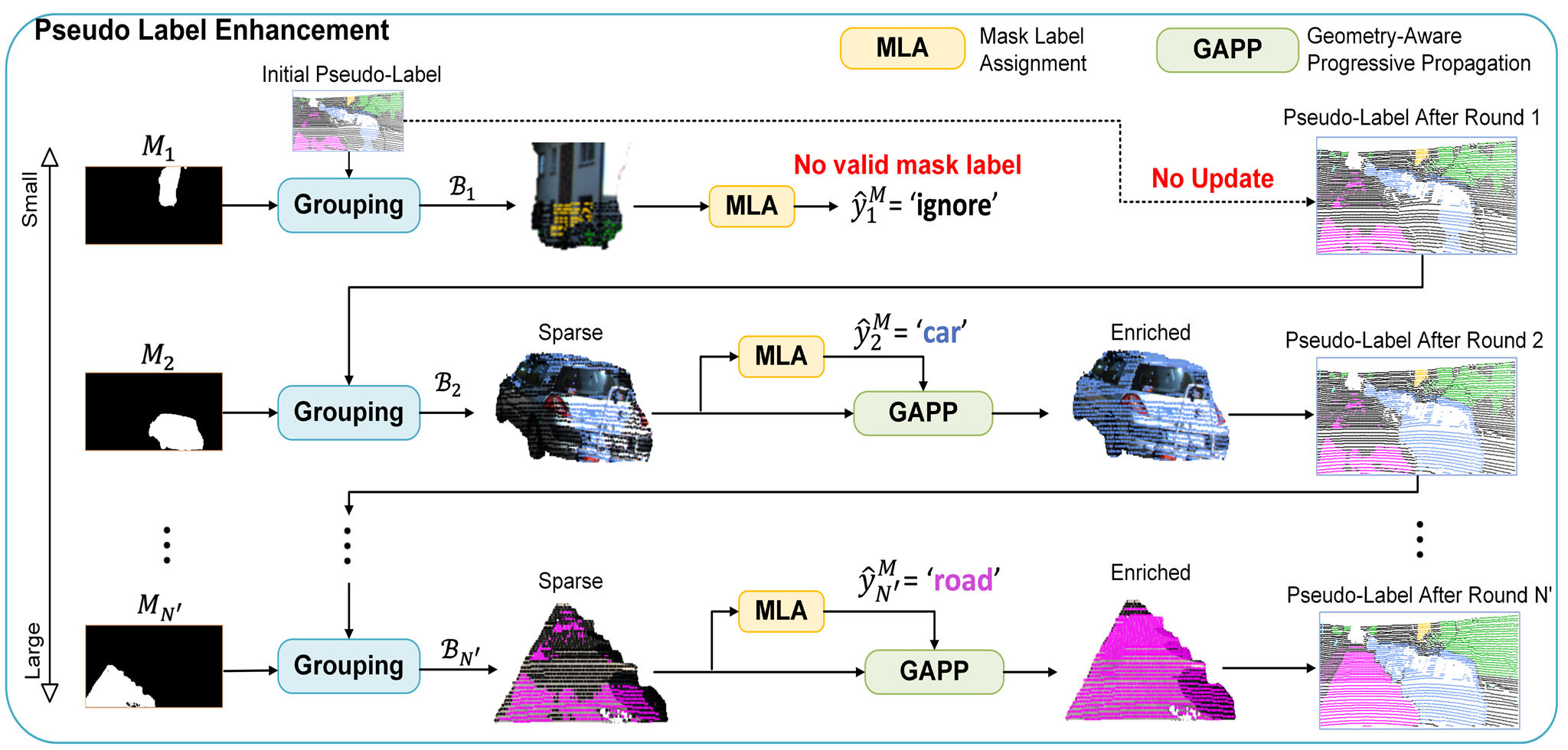}
\caption{\textbf{Visual explanation of the proposed method.} The SAM masks are sorted in descending order based on their area. For $M_1$, the pseudo-labels contain two main classes (yellow and green) and violate the purity constraint in Equation \ref{equ:rp}. Thus the mask label is not valid and there is no update to the pseudo-labels. For $M_2$ and $M_{N'}$, MLA generates valid mask labels and we apply GAPP to enhance the pseudo-labels within the mask. Then we update the pseudo-labels $\hat{y}$ and proceed to the next mask.}
\label{fig:structure}
\vspace{-0.2in}
\end{figure*}
}

\newcommand{\figProp}{
\begin{figure*}[t]
\centering
\includegraphics[width=0.93\linewidth]{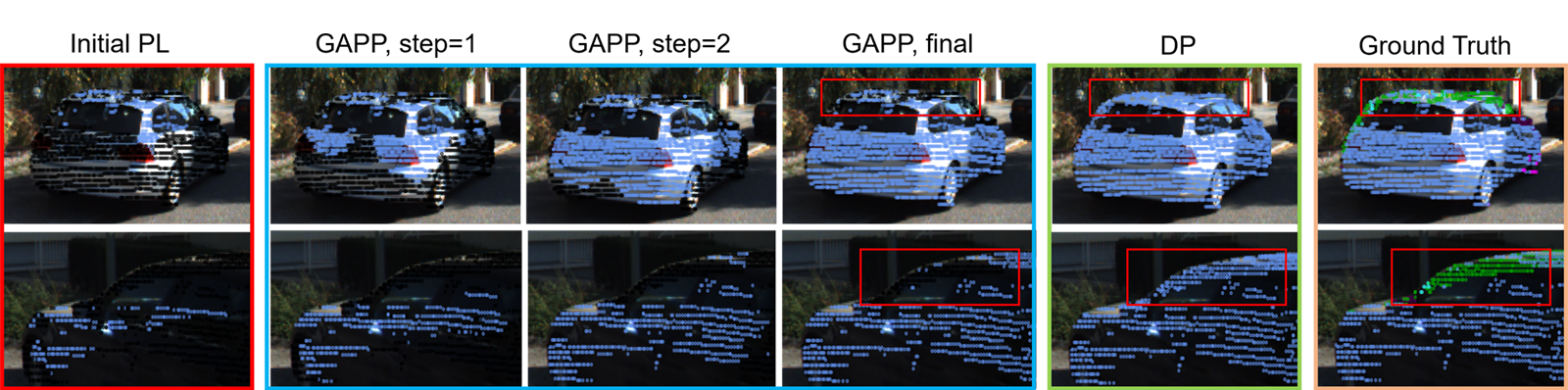}
\caption{\textbf{Visualization of the proposed GAPP}. Note the green points on the Ground Truth images (rightmost) belong to the background, not the car. The lidar paths of these green points are occluded in the camera image, thus they `appear' to be on the car due to the sensor location mismatch between the lidar and camera. Unlike Direct Propagation (DP), GAPP successfully avoids those incorrect mask labels assigned to the car.}
\label{fig:prop}
\vspace{-0.2in}
\end{figure*}
}

\newcommand{\figVis}{
\begin{figure*}[t]
\centering
\includegraphics[width=\linewidth]{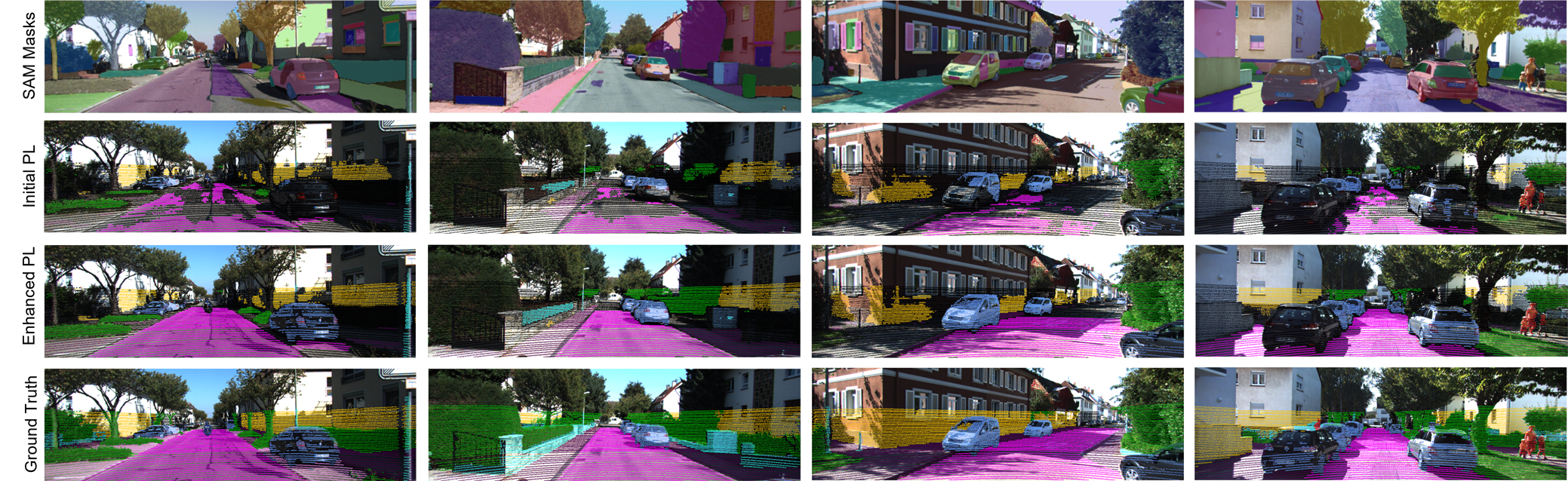}
\caption{Visualization of the SAM masks and enhanced pseudo-labels. The proposed method significantly improves the pseudo-label quality.}
\label{fig:visualization}
\vspace{-0.2in}
\end{figure*}
}

\newcommand{\algo}{
\begin{algorithm}[t]
\caption{Proposed Pseudo-Label Enhancement Method}
\label{alg:algo}
\begin{algorithmic}[1]
\Procedure{PLEnhancement}{$x^{2D}, x^{3D}, \hat{y}$, SAM} 
\State $\{M_j\}_{j=1}^{N'} \gets \text{SAM}(x^{2D})$ \Comment{Get SAM masks}
\State Sort the masks in descending order using area
\For{$j = 1$ to $N'$}
    \State Get $\mathcal{B}_j$ through Equation \ref{equ:getpoints}
    \State Get the dominant class $\tilde{c}$ through Equations \ref{equ:getpoints1}, \ref{equ:getclass}
    \State Get $R_{s}, R_{p}, R_{r}$ using Equation \ref{equ:rsize}, \ref{equ:rp}, \ref{equ:rpoint}
    \If{$R_{size} \leq \lambda_s, R_p \geq \lambda_p, R_r \geq \lambda_r$}
        \State $\hat{y}^M_j \gets \tilde{c}$  \Comment{Assign robust mask label}
        \State Calculate $d_{exp}$ using Equation \ref{equ:dist}
        \State Calculate $\mathcal{B}_{exp}$ using Equation \ref{equ:exp}
        \While{$\mathcal{B}_{exp} \neq \phi$}
            \State Update $\hat{y}$ using Equation \ref{equ:update}
            \State Recalculate $d_{exp}$ using Equation \ref{equ:dist}
            \State Recalculate $\mathcal{B}_{exp}$ using Equation \ref{equ:exp}
        \EndWhile
    \EndIf
\EndFor
\State \Return $\hat{y}$
\EndProcedure
\end{algorithmic}
\end{algorithm}
}

\newcommand{\tableMain}{
\begin{table*}[t]
\small
\centering
\caption{Main adaptation performance of our proposed method on mIoU ($\%$). The best results in UDA and SFDA are marked in bold and the performance improvement in 2D+3D is marked as green. }

\label{table:mainres}
\resizebox{\textwidth}{!}{
\begin{tabular}{cc|ccc|ccc|ccc}
\toprule
& \multirow{2}{*}{Method}   &  \multicolumn{3}{c|}{USA$\rightarrow$Singapore}  & \multicolumn{3}{c|}{A2D2$\rightarrow$SemKITTI}& \multicolumn{3}{c}{Singapore$\rightarrow$SemKITTI}   \\
           &  & {\footnotesize 2D}       & {\footnotesize 3D} & {\footnotesize 2D+3D}      & {\footnotesize 2D}    & {\footnotesize 3D}   & {\footnotesize 2D+3D}    & {\footnotesize 2D}       & {\footnotesize 3D}  & {\footnotesize 2D+3D} \\ \midrule

&  No Adaptation      & 47.27   & 45.47    &   49.13     & 32.8      & 32.59  & 39.49    & 24.69       &  29.71   & 32.49 \\ 
 \midrule
  \multirow{3}{*}{\begin{tabular}{@{}c@{}c@{}} UDA \end{tabular}} & xMUDA      & 56.18    &  50.85   &  58.86   & 41.12      & 44.06   & 46.97    &  33.17    &  34.73  &  37.17 \\
& $\text{xMUDA}_{pl}$     & 59.64  &  51.94  &  63.09   & 44.33      & 47.87   & 49.46    &    34.21   &  38.73  &  40.13 \\ 
& $\text{xMUDA}_{pl}$+ours    & \textbf{61.78} &   \textbf{53.96}   & \textbf{64.53} \rm \textcolor[RGB]{0,155,0}{(+1.44)}  &   \textbf{45.55}      & \textbf{51.21}   & \textbf{51.10}  \rm \textcolor[RGB]{0,155,0}{(+1.64)}       &    \textbf{37.92}   & \textbf{40.29}  &  \textbf{43.72} \rm \textcolor[RGB]{0,155,0}{(+3.59)} \\ \midrule
\multirow{2}{*}{SFDA} & SUMMIT      & 57.00   &  53.79   &  58.35  & 44.79      & 47.72   & 49.88    &      36.28    &  43.64  &   42.47  \\ 
& SUMMIT+ours      & \textbf{57.61}    &  \textbf{54.22}   &  \textbf{58.98} \rm \textcolor[RGB]{0,155,0}{(+0.63)} & \textbf{47.98}     & \textbf{49.64}   & \textbf{53.22} \rm \textcolor[RGB]{0,155,0}{(+3.34)} &     \textbf{41.54}    &  \textbf{45.74}  &  \textbf{47.34} \rm \textcolor[RGB]{0,155,0}{(+4.87)}  \\  \bottomrule

\end{tabular}
}
\vspace{-0.1in}
\end{table*}
}

\newcommand{\tableLabel}{
\begin{table}[t]
  \centering
  \caption{Comparison of pseudo-label accuracy (\%) and total correct pseudo-label increment (\%) for four classes under \textbf{A2D2 $\rightarrow$ SemKITTI}. }
  \label{table:label}
\resizebox{\linewidth}{!}{
    \begin{tabular}{c|ccc|cccc|cc}
    \toprule
    &  &   &   & \multicolumn{4}{c|}{Pseudo-Label Acc.}  &  &   \\
    & DP & MF in MLA  & GAPP  & Car & Truck &  Bike  & Person & Avg. Acc. &  Total Inc. \\
    \midrule
    \multirow{4}{*}{\begin{tabular}{@{}c@{}c@{}} xMUDA-2D \end{tabular}} & - & - & - & 89.77   & 9.52   & 72.20  & 55.82   & 56.83 & -    \\  
    & \ding{51} & - & - & 85.20  & 5.49  & 48.30  & 43.84    & 45.71 & +48.11\%    \\ 
    & \ding{51} & \ding{51} & - & 86.06  & 7.40  & 61.56  & 53.25   & 52.07 & +38.35\%    \\  
    & - & \ding{51} & \ding{51} & {86.92}  & {7.67} & {62.93}  & {54.83}   & {53.09} & +38.18\%    \\  
    \midrule
    \multirow{4}{*}{\begin{tabular}{@{}c@{}c@{}} xMUDA-3D \end{tabular}} & - & - & - &  96.95  & 40.89  & 87.09   & 17.45    & 60.60 & -    \\  
     & \ding{51} & - & - & 82.17  & 26.57   & 57.86   & 9.39  & 44.00 & +41.98\%    \\ 
     & \ding{51} & \ding{51} & - & 86.42  & 32.46   & 74.12   & 12.26   & 51.32 & +31.90\%    \\  
     & - & \ding{51} & \ding{51}& {94.57}  & {33.56}   & {81.63}  & {14.04}   & {55.95} & +30.39\%   \\  
    \midrule
    \multirow{4}{*}{\begin{tabular}{@{}c@{}c@{}} SUMMIT \end{tabular}} & - & - & - & 98.79  & 41.11  & 99.50   & 90.25    & 82.41 & -   \\  
     & \ding{51} & - & - & 81.35   & 27.04   & 77.25   & 43.40    & 57.26 & +159.5\%    \\ 
    &\ding{51} & \ding{51}& - & 84.14   & 34.94   & 83.21   & 60.33  & 65.66 & +124.6\%    \\  
    & - & \ding{51} & \ding{51} & {95.28}  & {37.44}   & {96.77}  & {75.31}  & {76.20} & +119.5\%  \\  
    \bottomrule
    \end{tabular}%
}
\end{table}
}

\newcommand{\tableAbl}{
\begin{table}[t]
\small
\centering
\caption{Ablation study of the effect of different components on mIoU ($\%$). Best numbers are marked in bold.}

\label{table:ablation}
\resizebox{\linewidth}{!}{
\begin{tabular}{c|ccc|ccc}
\toprule
Method  & DP & MF in MLA  & GAPP &  2D  & 3D & 2D+3D   \\  \midrule

  \multirow{4}{*}{\begin{tabular}{@{}c@{}c@{}} $\text{xMUDA}_{pl}$ \end{tabular}} &  - & - & - & 44.33    &  47.87   &  49.46  \\
& \ding{51} & - & - & 43.16  &  48.05   & 48.30 \\
& \ding{51} & \ding{51} & - & 45.46  &  50.22   & 50.62 \\
& - & \ding{51} & \ding{51} & \textbf{45.55}  &  \textbf{51.21}   & \textbf{51.10} \\ \midrule
\multirow{4}{*}{SUMMIT} & - & - & - &44.79   &  47.72  & 49.88   \\
& \ding{51} & - & - & 45.94  &  48.72   & 51.14 \\
& \ding{51} & \ding{51} & - & 47.37  &  48.52   & 52.14 \\
& - & \ding{51} & \ding{51} & \textbf{47.98}  &  \textbf{49.64}   & \textbf{53.22} \\ \bottomrule

\end{tabular}
}
\vspace{-0.2in}
\end{table}
}
\begin{abstract}
Multi-modal 3D semantic segmentation is vital for applications such as autonomous driving and virtual reality (VR). To effectively deploy these models in real-world scenarios, it is essential to employ cross-domain adaptation techniques that bridge the gap between training data and real-world data. Recently, self-training with pseudo-labels has emerged as a predominant method for cross-domain adaptation in multi-modal 3D semantic segmentation. However, generating reliable pseudo-labels necessitates stringent constraints, which often result in sparse pseudo-labels after pruning. This sparsity can potentially hinder performance improvement during the adaptation process.
We propose an image-guided pseudo-label enhancement approach that leverages the complementary 2D prior knowledge from the Segment Anything Model (SAM) to introduce more reliable pseudo-labels, thereby boosting domain adaptation performance. Specifically, given a 3D point cloud and the SAM masks from its paired image data, we collect all 3D points covered by each SAM mask that potentially belong to the same object. Then our method refines the pseudo-labels within each SAM mask in two steps. 
First, we determine the class label for each mask using majority voting and employ various constraints to filter out unreliable mask labels. Next, we introduce Geometry-Aware Progressive Propagation (GAPP) which propagates the mask label to all 3D points within the SAM mask while avoiding outliers caused by 2D-3D misalignment.
Experiments conducted across multiple datasets and domain adaptation scenarios demonstrate that our proposed method significantly increases the quantity of high-quality pseudo-labels and enhances the adaptation performance over baseline methods.
\end{abstract}
\section{INTRODUCTION}
In recent years, 3D semantic segmentation has emerged as a pivotal task in 3D scene understanding, essential for applications such as autonomous driving  \cite{wu2018squeezeseg, xu2021rpvnet} and augmented/virtual reality (AR/VR) \cite{choy20194d, thomas2019kpconv, qi2017pointnet++}. Recently, driven by the new multi-modal datasets \cite{caesar2020nuscenes, behley2019semantickitti}, the integration of image data has been increasingly adopted to enhance the accuracy of 3D semantic segmentation as it provides complementary 2D information such as rich texture and color details, which supplements the geometric information from 3D point clouds \cite{su2018splatnet, meyer2019sensor}. 

However, similar to other perception tasks, 3D semantic segmentation can suffer from domain shifts between the training and the real-world testing environments, necessitating the use of domain adaptation techniques. Domain adaptation aims to bridge this domain gap and has been a significant area of research for both 2D and 3D semantic segmentation \cite{hoffman2018cycada, li2019bidirectional, vu2019advent, wu2019squeezesegv2, yi2021complete}. Recently, such focus has extended to multi-modal 3D semantic segmentation and has shown impressive performance across various adaptation settings \cite{jaritz2020xmuda, cao2024mopa, shin2022mm, simons2023summit}. Among these existing methods, self-training using pseudo-labels has proven to be a crucial component and several pseudo-label generation approaches have been proposed such as thresholding \cite{jaritz2020xmuda} or modality agreement \cite{shin2022mm, simons2023summit}. However, the generated pseudo-labels tend to be sparse (as shown in Fig. \ref{fig:front}) and often limit the overall adaptation performance due to its insufficient coverage of the target data. 

\figFront

Recently, the Segment Anything Model (SAM) \cite{kirillov2023segment, ravi2024sam} has gathered significant attention. Trained over 1 billion masks, SAM has exhibited remarkable zero-shot segmentation capabilities by generating class-agnostic segmentation masks with proper prompts. This zero-shot ability has proven valuable for various applications \cite{huang2023push, ma2024segment, huang2024segment, chen2023segment}. Although SAM is not inherently designed for 3D point clouds, its versatility has been extended to multi-modal settings in recent works \cite{cao2024mopa, peng2023sam, liu2024segment}. 

Inspired by recent advancements in leveraging the zero-shot segmentation capability of SAM, we propose to employ such 2D prior knowledge in the multi-modal setting to enhance sparse 3D pseudo-labels and hence boost the domain adaptation performance. Given a 3D point cloud, the SAM masks generated from the paired 2D image data can effectively group 3D points that belong to the same object by exploiting the 3D-2D correspondence between 3D points and the camera plane. With this grouping information, we design a two-step mask-wise pseudo-label enhancement framework to generate additional reliable pseudo-labels within each SAM mask. 
Specifically, for each SAM mask associated with multiple 3D pseudo-labels for its covered points, we first identify the class label of the entire mask through majority voting. To alleviate the effect of inherent pseudo-label noise, we introduce various constraints on the mask area and the distribution of pseudo-labels to filter out unreliable mask labels. Secondly, we aim to propagate the mask label to all points within the mask that lack pseudo-labels. To avoid assigning the mask label to outlier points that are incorrectly projected to the object due to 2D-3D misalignment \cite{an2020geometric}, we propose Geometry-Aware Progressive Propagation (GAPP) where the mask label is propagated only to nearby points in the 3D space in each round, thus eliminating outlier points that lack a connection to the object.   
We evaluate the proposed method on multiple datasets and two adaptation tasks: unsupervised and source-free domain adaptation. Experimental results demonstrate that our method effectively increases the number of high-quality pseudo-labels and significantly improves the adaptation performance.

Our contributions can be summarized as follows:
\begin{itemize}
   \item We introduce a novel pseudo-label enhancement method for multi-modal 3D semantic segmentation domain adaptation tasks using 2D SAM masks. 
   \item We propose a series of mask label filtering constraints to ensure robust mask label generation and a geometry-aware propagation strategy GAPP to tackle the 2D-3D misalignment issue. 
   \item The proposed method is evaluated on two adaptation tasks and three adaptation scenarios using multiple datasets, consistently demonstrating improvement over existing domain adaptation methods.
\end{itemize}
\section{Related Works}

\subsection{Unsupervised Domain Adaptation (UDA) For Semantic Segmentation}

\textbf{Uni-modal:} UDA aims to address the challenge of domain shift, particularly in scenarios where no labeled data is available in the target domain. It has been extensively studied in the context of uni-modal semantic segmentation tasks for both 2D and 3D scenarios. For 2D semantic segmentation, one common approach is to align the distribution of featuresoutputs between the source and target domains \cite{tsai2018learning, hoffman2018cycada, hong2018conditional, vu2019advent}. Alternative approaches  pseudo-labeling techniques to populate missing labels to reduce the domain gap \cite{li2022class, mei2020instance, pan2020unsupervised, zhao2023learning}.
Similarly, several notable contributions have been made to 3D semantic segmentation. Wu et al. \cite{wu2019squeezesegv2} introduced activation correlation alignment and progressive domain calibration to perform UDA between simulated and real-world LiDAR data. Yi et al. \cite{yi2021complete} proposed the `Complete \& Label' framework, which models the underlying 3D surface priors to allow the transfer of semantic labels between different LiDAR sensors. Additionally, Saltori et al. \cite{saltori2022cosmix} introduced CoSMix, which alleviates domain shift through a point cloud mixing strategy.

\textbf{Multi-modal:} Recently, the growth of multi-modal datasets \cite{behley2019semantickitti, caesar2020nuscenes, geyer2020a2d2} containing both LiDAR measurements and corresponding images, along with advancements in multi-modal methodologies \cite{bai2022transfusion, meyer2019sensor, yang2022efficient}, has increased the interest in UDA within multi-modal settings. Jaritz et al. \cite{jaritz2020xmuda} introduced the first UDA framework for multi-modal scenarios, xMUDA, which incorporates two additional head networks to enforce cross-modal similarity. Building on xMUDA, Peng et al. \cite{peng2021sparse} introduced DsCML to enhance information interaction between different modalities for UDA. Similarly, Xing et al. \cite{xing2023cross} proposed an attention module to improve the integration of 2D features with 3D points. Additionally, source-free domain adaptation (SFDA) \cite{simons2023summit} and test-time domain adaptation (TTA) \cite{shin2022mm, cao2024reliable} have been studied as specialized cases within UDA.

\subsection{Segment Anything Model (SAM)}

Recently, SAM \cite{peng2023sam, ravi2024sam} has gathered significant attention as a versatile segmentation model, trained on the extensive SA-1B dataset with over 1 billion masks derived from 11 million images. SAM demonstrates remarkable zero-shot transfer capability and this has catalyzed the application of SAM across numerous domains, such as medical image segmentation \cite{huang2023push, huang2024segment}, weakly-supervised image segmentation \cite{chen2023segment, he2024weakly}, object detection \cite{yang2023sam3d}, and tracking \cite{yang2023track, cheng2023segment}. Such concepts have also been extended to 3D scenarios \cite{liu2024segment}. Beyond leveraging SAM's zero-shot abilities, several studies have proposed fine-tuning the SAM for various downstream tasks \cite{ma2024segment, zhang2023personalize, wu2023medical}.

The zero-shot capability of SAM has recently been applied to multi-modal 3D semantic segmentation domain adaptation. For instance, Peng et al. \cite{peng2023sam} proposed a technique to map both 2D and 3D features to SAM features to bridge the domain gap. Similarly, Cao et al. \cite{cao2024mopa} utilize SAM masks to impose a pixel-wise consistency loss, thus encouraging consistent predictions within each mask. These existing methods aim to improve the UDA performance by refining the alignment between the source and target models using SAM, while our method focuses on employing SAM masks to enhance the quality of 3D pseudo-labels for self-training. Furthermore, existing methods typically assume the constant presence of the source model, which hampers their generalizability to SFDA or TTA scenarios when the source model is absent. On the contrary, our method is flexible and suitable for diverse adaptation scenarios where pseudo-labeling is possible, regardless of the source model.
\section{Proposed Method}

\figStructure

\subsection{Preliminary of Self-Training in UDA}
For UDA in the multi-modal 3D semantic segmentation task, we utilize a source domain dataset $\mathcal{D}_s$ which includes 3D LiDAR point clouds $x_s^{3D} \in \mathcal{X}_s^{3D}$, their corresponding 2D images $x_s^{2D} \in \mathcal{X}_s^{2D}$, and ground truth labels $y_s \in \mathcal{Y}_s$ for each point in $x_s^{3D}$. In contrast, the target domain consists of only the target point clouds $x_t^{3D} \in \mathcal{X}_t^{3D}$ and their paired 2D images $x_t^{2D} \in \mathcal{X}_t^{2D}$, without any label information. Recent cross-domain adaptation methods employ pseudo-labels $\hat{y}_t$ to facilitate the adaptation process through self-training. Generally, the pseudo-label based self-training objective can be divided into two main components: 
\begin{equation}
    \mathcal{L} = \mathcal{L}_s(x^{2D}_s, x^{3D}_s, y_s) + \lambda \mathcal{L}_t(x^{2D}_t, x^{3D}_t, \hat{y}_t),
    \label{equ:overview}
\end{equation}
where $\mathcal{L}_s$ represents the supervised loss on the source domain, $\mathcal{L}_t$ denotes the unsupervised loss on the target domain, and $\lambda$ is a weighting factor. In the particular case of source-free domain adaptation, the adaptation process is conducted without access to the source data, resulting in Equation \ref{equ:overview} containing only the target loss $\mathcal{L}_t$. Obviously, the quality of the pseudo-labels $\hat{y}_t$ is crucial for enhancing adaptation performance in this context. To generate high-quality 3D pseudo-labels, several methods have been proposed to filter/prune out unreliable labels, including thresholding \cite{jaritz2020xmuda}, entropy weighting \cite{shin2022mm}, and cross-modal agreement \cite{shin2022mm, simons2023summit}. However, as noted in \cite{simons2023summit}, reliable pseudo-labels after filtering/pruning are often sparse, especially in cases with large domain gaps. To address this challenge, we aim to develop a method to enhance the density of reliable pseudo-labels and boost the adaptation performance.

\figProp

\subsection{Method Overview}
The overview of the proposed pseudo-label enhancement method is summarized in Fig. \ref{fig:structure}. For a simpler notation, let $x^{3D} \in \mathbb{R}^{3\times N}$ denote the target LiDAR data and $x^{2D} \in \mathbb{R}^{3\times H \times W}$ denote the paired image data, where $H$ and $W$ defines the image size and $N$ is the total number of points. The 3D point cloud $x^{3D}$ can be further represented as $x^{3D} = \{o_k\}_{k=1}^{N}$, where $o_k \in \mathbb{R}^3$ denotes the 3D coordinate of the $k$-th point. Let $\hat{y} = \{\hat{y}_k\}_{k=1}^{N} $ represent the pseudo-labels for $x^{3D}$, with $\hat{y}_k$ denoting the pseudo-label for point $o_k$. We assume that $\hat{y}$ has been initialized using a certain pseudo-label generation strategy as in \cite{jaritz2020xmuda, simons2023summit}. For the 2D image $x^{2D}$, we extract a set of masks from SAM, denoted as $\mathcal{M} = \{M_j\}_{j=1}^{N'}$, where $N'$ is the total number of masks in $x^{2D}$. These masks are sorted in descending order based on their area. Our method iteratively updates the pseudo-labels $\hat{y}$, beginning with the smallest mask $M_1$ and proceeding to the largest mask $M_{N'}$. We prioritize smaller masks for updates as each mask is more likely to correspond to an individual object sharing a common label within the mask.

For the $j$-th mask $M_j$, we first identify the set of 3D points whose 2D projections lie within $M_j$, which is expressed as:
\begin{equation}
\mathcal{B}_{j} = \{1 \leq k \leq N\ | \;\;  \text{proj}(o_k) \in M_j \},
\label{equ:getpoints}
\end{equation}
where $\text{proj}(.)$ represents the 3D-2D projection function. After identifying the corresponding 3D points for mask $M_j$, we first perform Mask Label Assignment (MLA, details in Section \ref{MLA}) to generate its mask label $\hat{y}^M_{j}$. If a valid mask label $\hat{y}^M_{j}$ is not obtained (i.e., \textit{ignore} case for $M_{1}$ in Fig. \ref{fig:structure}), we skip updating the pseudo-labels $\hat{y}$ and proceed directly to the next mask (e.g., from $M_1$ to $M_2$ in Fig. \ref{fig:structure}). If $\hat{y}^M_{j}$ is valid (e.g., $\hat{y}^M_{2}$ and $\hat{y}^M_{N'}$ for $M_2$ and $M_{N'}$ in Fig. \ref{fig:structure}), we apply Geometry-Aware Progressive Propagation (GAPP, Section \ref{GAPP}) to propagate the mask label to the remaining points lacking valid pseudo-labels, while excluding outlier points. After applying GAPP, we update the pseudo-labels $\hat{y}$ and move on to the next mask $M_{j+1}$. The following sections provide the details of the MLA process and the GAPP module.

\subsection{Mask Label Assignment} \label{MLA}
The goal of MLA is to assign each mask $M_j$ a class label $\hat{y}^{M}_j$ based on the valid pseudo-labels in $\mathcal{B}_{j}$. A straightforward approach to achieve this is by majority voting, wherein the most frequent pseudo-label within $\mathcal{B}_{j}$ is chosen. Specifically, the points classified as class $c$ are denoted as:
\begin{equation}
\mathcal{B}_{j}^c = \{k | k \in \mathcal{B}_{j}, \hat{y}_k = c\}.
\label{equ:getpoints1}
\end{equation}
Then, the dominant class $\tilde{c}$ can be determined by:
\begin{equation}
\tilde{c} = \argmax_c |\mathcal{B}_{j}^c|.
\label{equ:getclass}
\end{equation}

However, the pseudo-label $\tilde{c}$ may not accurately represent the mask $M_j$. For instance, a large mask might encompass multiple objects, and assigning a single label would lead to incorrect labeling. Additionally, noisy pseudo-labels can further undermine the accuracy of the mask label. To address these challenges, we introduce Mask Filtering (MF) which contains three constraints based on the mask size, label purity, and label representativity. Only masks that satisfy all three constraints are assigned a valid label. 

\textbf{Mask Size}: To ensure masks are relatively small, we impose a constraint based on the ratio between the mask size and the image size, given by the equation:
\begin{equation}
R_{s} = \frac{\Omega(M_j)}{HW} \leq \lambda_s,
\label{equ:rsize}
\end{equation}
where $\Omega(.)$ is the mask area and $\lambda_s$ is a threshold parameter. 

\textbf{Label Purity}: To ensure that $\tilde{c}$ is the only dominant class within the mask, we evaluate the purity by examining the ratio of the number of points labeled with $\tilde{c}$ over the number of all valid labels. Let $\mathcal{B}_j^{ignore}$ denote the points in $M_j$ that does not have a valid label. Then the purity constraint can be defined as follows:
\begin{equation}
R_{p} = \frac{|\mathcal{B}_j^{\tilde{c}}|}{|\mathcal{B}_j|-|\mathcal{B}_j^{ignore}|} \geq \lambda_p,
\label{equ:rp}
\end{equation}
where $\lambda_p$ is a parameter limiting the lowest purity.

\textbf{Label Representativity}: We also require that the pseudo-label of class $\tilde{c}$ are representative for the entire mask. To ensure this, the number of points that belong to the dominant class $\tilde{c}$ should not be too small compared to all the points within the mask. This can be expressed as:
\begin{equation}
R_{r} = \frac{|\mathcal{B}_j^{\tilde{c}}|}{|\mathcal{B}_j|} \geq \lambda_r, 
\label{equ:rpoint}
\end{equation}
where $\lambda_r$ is a threshold parameter that controls the ratio.

Finally, the MLA process can be summarized as:
\begin{equation}
\hat{y}_j^{M} = 
\begin{cases}
\tilde{c} & \text{if Equations \ref{equ:rsize}, \ref{equ:rp}, \ref{equ:rpoint} hold,} \\
  \textit{ignore} & \text{otherwise.}
\end{cases}
\end{equation}

\subsection{Geometric-Aware Progressive Propagation (GAPP)} \label{GAPP}

\algo
\tableMain
\tableLabel

Once we have a valid mask label, the next objective is to propagate this label from $\mathcal{B}_j^{\tilde{c}}$ to all points lacking a valid pseudo-label (i.e., $\mathcal{B}_j^{ignore}$). A straightforward approach to achieve this is Direct Propagation (DP), where the mask label is assigned to every point in $\mathcal{B}_j^{ignore}$ as follows:
\begin{equation}
\hat{y}_k = \hat{y}^M_j, \;\; \forall k \in \mathcal{B}_{j}^{ignore}.
\end{equation}

In a multi-modal setting, a significant challenge is the 2D-3D alignment issue \cite{an2020geometric}, where some 3D points are incorrectly projected onto objects due to occlusion (e.g., background points seen by the lidar are occluded by an object in the camera image). As illustrated by the ground truth labels in Fig. \ref{fig:prop}, several background points (e.g., nature) are erroneously projected onto foreground objects (e.g., cars). Consequently, DP inevitably introduces additional pseudo-label errors. Inspired by distance-based point cloud outlier removal algorithms \cite{zhang2009new, ning2018efficient}, we leverage the 3D information of points and propose a progressive propagation algorithm to eliminate outlier background points that lack a connection to $\mathcal{B}_j^{\tilde{c}}$. In each iteration, we only propagate the mask label $\hat{y}^M_j$ to `nearby' points with a relatively small exploration distance $d_{exp}$. To define $d_{exp}$, we first define the minimum pair-wise distance between point $o_i$ and the points in $\mathcal{B}_j^{\tilde{c}}$ as: 
\begin{equation}
d_i = \min_{k \neq i, k \in \mathcal{B}_j^{\tilde{c}}} || o_i - o_k ||_2.
\end{equation}
Then, the exploration distance $d_{exp}$ is defined as the maximum pair-wise distance within $\mathcal{B}_j^{\tilde{c}}$, which is:
\begin{equation}
d_{exp} = \max_{i \in \mathcal{B}_j^{\tilde{c}}} d_i. 
\label{equ:dist}
\end{equation}
Next, with the exploration distance, the `nearby' points to propagate the label, denoted as $\mathcal{B}_{exp}$, can be determined by:
\begin{equation}
\mathcal{B}_{exp} = \{ k \in \mathcal{B}_j^{ignore} | d_k \leq \beta d_{exp}\},
\label{equ:exp}
\end{equation}
where $\beta$ is a scaling factor of this exploration distance. Then, the pseudo-labels $\hat{y}$ can be updated through:
\begin{equation}
\hat{y}_k = \hat{y}^M_j, \;\; \forall k \in \mathcal{B}_{exp}.
\label{equ:update}
\end{equation}
After updating $\hat{y}$, we recalculate the sets $\mathcal{B}_j^{\tilde{c}}$ and $\mathcal{B}_j^{ignore}$, and repeat this process until the set of points to explore, $\mathcal{B}_{exp}$, is empty. This iterative procedure is illustrated in Fig. \ref{fig:prop}. Starting from the initial pseudo-labels, GAPP progressively propagates the mask label across the entire object while leaving out the outliers that are not connected to the object. The proposed pseudo-label enhancement algorithm is summarized in Algorithm \ref{alg:algo}.

\section{Experiments}

\tableAbl
\subsection{Benchmarks}
Following the methodology described in \cite{jaritz2020xmuda, simons2023summit}, we evaluate our method across three different benchmarks: (1) \textbf{USA-to-Singapore}, (2) \textbf{A2D2-to-SemanticKITTI}, and (3) \textbf{Singapore-to-SemanticKITTI}. The first scenario (1) uses the NuScenes dataset \cite{caesar2020nuscenes} for both the source and target domains which are separated by the location of scenes. The point clouds and images collected in the USA serve as the source domain, while those collected in Singapore are used as the target domain. The primary domain gap arises from the differing infrastructure between the two countries, and we create a 5-class segmentation benchmark based on these differences. In the second scenario (2), we use A2D2 dataset \cite{geyer2020a2d2} as the source domain and SemanticKITTI \cite{behley2019semantickitti} dataset as the target domain. Following the class-mapping strategies outlined in \cite{jaritz2020xmuda, simons2023summit}, we establish a 10-class semantic segmentation benchmark. The third scenario (3) uses the Singapore dataset from NuScenes as the source domain and the SemanticKITTI dataset as the target domain. We follow \cite{simons2023summit} and formulate a 10-class semantic segmentation benchmark using the mapping in NuScenes-LiDARSeg dataset. The scenario (2) and (3) specifically address dataset-to-dataset domain gap/adaptation, primarily focusing on differences in sensor settings (e.g., number of LiDAR beams, image resolution) and mounting positions. It is important to note that we only utilize the 3D points with valid projection onto the camera plane.

\figVis

\subsection{Implementation Details}

We consider two adaptation tasks for multi-modal 3D semantic segmentation: unsupervised domain adaptation (UDA) and source-free domain adaptation (SFDA). We select \textbf{xMUDA}\cite{jaritz2020xmuda} and \textbf{SUMMIT}\cite{simons2023summit} as our baselines, as both methods utilize pseudo-labeling during their adaptation process. Our experiments are implemented in PyTorch and we strictly follow the network structures in \cite{jaritz2020xmuda} and \cite{simons2023summit}. Specifically, we use ResNet34 \cite{he2016deep} as the 2D backbone and the Sparse Convolutional Network UNet \cite{graham20183d} as the 3D backbone. For the UDA setting, we follow the training strategy of xMUDA \cite{jaritz2020xmuda} with the only difference of replacing their pseudo-labels with our enhanced ones. Note that xMUDA produces pseudo-labels from both 2D and 3D modalities, and we enhance both of them. For the SFDA setting, we follow the non-crossover training strategy described in SUMMIT \cite{simons2023summit} to train the models on the source domain. Then we adopt the Agreement Filtering strategy to generate the initial pseudo-labels and apply our method for additional pseudo-label enhancement. Subsequently, we update the trained model using the enhanced pseudo-labels for 50000 iterations with a learning rate of $1e-6$ for scenario (1) and $1e-4$ for scenarios (2) and (3). The Adam optimizer \cite{kingma2014adam}  is used with a batch size of 8 for all experiments. All training and testing are conducted using one NVIDIA A40 48GB GPU. For our pseudo-label enhancement algorithm, we use the pre-trained SAM-ViT-H checkpoint and the default settings of SAM to generate 2D masks. Consistent performance improvement was observed with the hyperparameters $\lambda_s=0.2$, $\lambda_p=0.8$, $\lambda_r=0.1$, and $\beta=2$ across all experiments.

\subsection{Main Results}

Table \ref{table:mainres} showcases the performance improvement on 3D semantic segmentation achieved by our proposed pseudo-label enhancement approach. We report the mean Intersection over Union (mIoU) for both modalities 2D and 3D, as well as the average across the softmax output of both modalities, denoted as 2D+3D. Under the UDA setting, the results demonstrate that xMUDA significantly outperforms the baseline without any adaptation. With the inclusion of psuedo-labels, $\text{xMUDA}_{pl}$ exhibits further improvements, consistent with the findings in \cite{jaritz2020xmuda}. Notably, when our proposed method is used to enhance pseudo-labels, xMUDA's performance improves across all three adaptation scenarios. Specifically, in the USA-to-Singapore scenario, where the domain gap is relatively small, we observe an improvement of up to $3.6\%$ for single modality and $2.28\%$ on 2D+3D. In the other two dataset-to-dataset adaptation scenarios, our method's improvements are more substantial, reaching up to $10.84\%$ for single modality and $8.95\%$ for the combined 2D+3D mIoU.
Under the SFDA setting, a similar trend is observed. The mIoU for 2D, 3D, and 2D+3D all show improvements. For instance, in the relatively small domain dap setting of USA-to-Singapore, the mIoU for 2D+3D improves by $0.87\%$. In other settings, the improvements are more significant, reaching $14.5\%$ for single modality and $11.47\%$ for both 2D and 3D modalities. These results demonstrate the effectiveness of our proposed method in enhancing pseudo-labels for both UDA and SFDA tasks.

Next, we perform an ablation study on the pseudo-label statistics and the mIoU performance when incorporating different components of our proposed method, as shown in Table \ref{table:label} and \ref{table:ablation}. All the experiments are conducted under the A2D2-to-SemanticKITTI scenario. When we perform pseudo-label enhancement using only DP, Table \ref{table:label} shows that the number of correct pseudo-labels increases by over $40\%$ for xMUDA and approximately $160\%$ for SUMMIT. However, due to pseudo-label noise and 2D-3D misalignment, this straightforward enhancement significantly reduces pseudo-label accuracy, as reflected in Table \ref{table:label}. Consequently, the improvement in mIoU, as shown in Table \ref{table:ablation}, is not significant. In fact, for xMUDA, mIoU performance decreases from 49.46 to 48.30. Next, when we add the MF criterion, although the number of correct pseudo-labels decreases slightly as some correct labels are filtered by MF as well, the pseudo-label accuracy significantly increases, leading to noticeable mIoU improvements in Table \ref{table:ablation} over the baseline without any pseudo-label enhancement. Lastly, incorporating the GAPP model to address the 2D-3D misalignment issue further enhances pseudo-label accuracy while maintaining a similar number of correct pseudo-labels.  This results in a pseudo-label accuracy comparable to the original pseudo-labels but with more than $30\%$ additional correct pseudo-labels for xMUDA and approximately $120\%$ more for SUMMIT. As the quality of pseudo-labels improves, the mIoU performance also increases, achieving the best results.

Finally, we include visualizations of the enhanced pseudo-labels in Fig. \ref{fig:visualization}. It shows the initial pseudo-labels are relatively sparse and lack comprehensive coverage of the entire scene. With our proposed algorithm, the resulting pseudo-labels are much denser, particularly for cars and roads, and it exhibits high accuracy by correctly avoiding mislabels for background points.

\section{Conclusion}

In this paper, we propose a novel pseudo-label enhancement approach for multi-modal 3D semantic segmentation domain adaptation, guided by 2D SAM masks. The proposed method refines pseudo-labels within each SAM mask using MLA and GAPP, ensuring robustness against pseudo-label noise and the 2D-3D misalignment issue. Experiments conducted on multiple domain adaptation tasks and scenarios demonstrate that the proposed algorithm successfully produces denser pseudo-labels with similar/higher pseudo-label accuracy. The enhanced pseudo-labels significantly improve the adaptation performance.

\newpage

\bibliographystyle{./IEEEtran}
\bibliography{./IEEEabrv,./IEEEexample}

\end{document}